\documentclass[10pt,twocolumn,letterpaper]{article}

\usepackage[pagenumbers]{cvpr} 

\usepackage{graphicx}
\usepackage{amsmath}
\usepackage{amssymb}
\usepackage{booktabs}
\usepackage[pagebackref,breaklinks,colorlinks]{hyperref}
\usepackage{xcolor}         
\usepackage{enumitem}
\usepackage{pifont}
\usepackage{tablefootnote}

\DeclareMathOperator*{\argmax}{arg\,max}

\newcommand{\cmark}{\ding{51}}%
\newcommand{\xmark}{\ding{55}}%

\usepackage[capitalize]{cleveref}
\crefname{section}{Sec.}{Secs.}
\Crefname{section}{Section}{Sections}
\Crefname{table}{Table}{Tables}
\crefname{table}{Tab.}{Tabs.}

\def\cvprPaperID{6792} 
\def\confName{CVPR}
\def\confYear{2023}

\newcommand*{\modelname}{Video-Audio Separation through Text\xspace}

\newcommand*{\modelabb}{VAST\xspace}

\begin{document}

%
\title{Language-Guided Audio-Visual Source Separation via Trimodal Consistency}


\author{Reuben Tan$^{1}$ \ \ \ \ Arijit Ray$^{1}$ \ \ \ \ Andrea Burns$^{1}$ \ \ \ \ Bryan A. Plummer$^{1}$ \ \ \ \  Justin Salamon $^{2}$ \ \ \ \  \\
Oriol Nieto$^{2}$ \ \ \ \ Bryan Russell$^{2}$ \ \ \ \ Kate Saenko$^{1,3}$ \\
$^{1}$Boston University, $^{2}$Adobe Research, $^{3}$MIT-IBM Watson AI Lab, IBM Research \\
{\tt \small \{rxtan, aburns4, array, bplum, saenko\}@bu.edu}, {\tt \small \{salamon, onieto, brussell\}@adobe.com} \\
} 
\maketitle

\begin{abstract}
   We propose a self-supervised approach for learning to perform audio source separation in videos based on natural language queries, using only unlabeled video and audio pairs as training data. A key challenge in this task is learning to associate the linguistic description of a sound-emitting object to its visual features and the corresponding components of the audio waveform, all without access to annotations during training. To overcome this challenge, we adapt off-the-shelf vision-language foundation models to provide pseudo-target supervision via two novel loss functions and encourage a stronger alignment between the audio, visual and natural language modalities.  
   During inference, our approach can separate sounds given text, video and audio input, or given text and audio input alone. We demonstrate the effectiveness of our self-supervised approach on three audio-visual separation datasets, including MUSIC, SOLOS and AudioSet, where we outperform state-of-the-art strongly supervised approaches despite not using object detectors or text labels during training. Our project page including publicly available code can be found at \href{https://cs-people.bu.edu/rxtan/projects/VAST}{https://cs-people.bu.edu/rxtan/projects/VAST}. 
   
\end{abstract}

\begin{figure}
    \centering
    \includegraphics[width=0.47\textwidth]{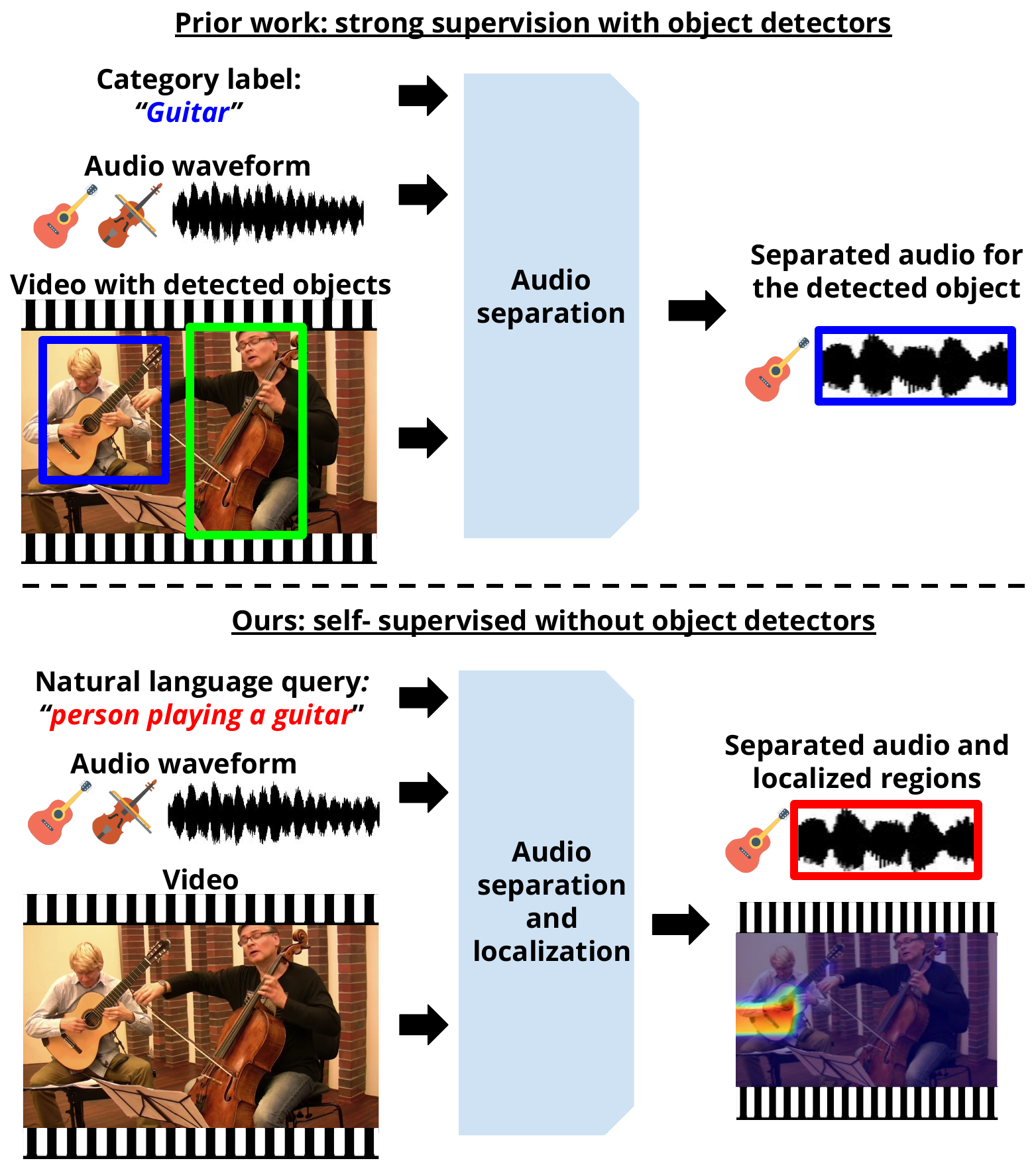}
    \caption{We propose to separate and localize audio sources based on a natural language query, by learning to align the modalities on completely unlabeled videos.
    In comparison, prior audio-visual sound separation approaches require object label supervision.}
    \label{fig:latent_motiv}
    \vspace{-10pt}
\end{figure}

\section{Introduction}
Our everyday audiovisual world is composed of many visible sound sources, often with multiple sources layering on top of one another. For example, consider the video of the guitar and cello musicians playing together in Fig.~\ref{fig:latent_motiv}. The two instruments have distinct timbres, and the musicians play non-unison, but complementary melodies. Despite hearing both instruments simultaneously, humans have an innate ability to 
identify and isolate the melody of a single source object.
In this paper, we define the corresponding machine task as follows: given a natural language query that selects a sounding object, such as ``person playing a guitar'', separate its sound source from the input audio waveform and localize it in the input video, without any supervision. 



This task is challenging. First, there is no approach for associating the linguistic description of a sound-emitting object to its visual features and the corresponding components of the audio waveform without access to annotations during training. Existing audio-visual methods \cite{zhao2018sound,gao2019co,chatterjee2021visual} do not generalize to natural language queries due to their dependence on discrete object class labels. Second, an ideal solution would jointly identify and localize sound-emitting objects in videos as well as separate the corresponding components in the audio waveform without strong supervision. Although prior audio-visual work has demonstrated the benefits of aligning relevant object regions in the video with their corresponding sounds  \cite{gao2019co,chatterjee2021visual}, these approaches require strong supervision including object label and bounding box annotations (see Fig.~\ref{fig:latent_motiv} top). Overcoming these challenges would enable important downstream applications including holistic video understanding \cite{seo2022end}, embodied AI \cite{chen2019audio}, and bidirectional audio-to-video retrieval \cite{wu2022wav2clip}.


To address these challenges, we make the following contributions. First, we propose \modelname (\modelabb), a self-supervised approach that leverages large vision-language ``foundation'' models~\cite{radford2021learning,jia2021scaling} to provide pseudo-supervision for learning the alignment between the three modalities: audio, video and natural language. 
Our key insight is to learn a strong \emph{transitive} relation from audio to natural language using vision as an intermediary modality, while preserving the alignment between the visual and natural language modalities embodied by the foundation models.  However, just using the visual representations of these foundation models in existing AV separation approaches does not preserve the transitive relationships between the three modalities (Sec.~\ref{sec:results}).

Our second contribution introduces two novel multimodal alignment objectives that encourage the learnt audio representations to encode the semantics of captions and infer the latent transitive relation between the three modalities.  While natural language can express a large and varied range of visual concepts for audio separation in videos, the absence of captions in unlabeled videos during training poses a significant challenge in our self-supervised formulation. To learn the transitive alignment, we adapt a foundation model to extract \textit{latent captions} from unlabeled videos.
Intuitively, the latent captions are representations that express the visual concepts present in the videos.  Third, we introduce a Multiple Instance Learning formulation to learn to perform audio separation at the video \emph{region} level since we do not have prior information on relevant objects or their locations in the videos during training.

Finally, we demonstrate the effectiveness of our proposed \modelabb approach through extensive evaluations on the audio source separation task on the SOLOS~\cite{montesinos2020solos}, MUSIC~\cite{zhao2018sound}, and AudioSet~\cite{gemmeke2017audio} datasets. We show that our self-supervised approach outperforms strongly-supervised state-of-the-art approaches without using labels during training by leveraging the capability of vision-language foundation models. More importantly, we demonstrate that \modelabb learns to use language queries for audio separation despite not training with ground-truth language supervision.

\section{Related Work}

\noindent\textbf{Audio-only source separation.} The goal of audio-only source separation is to use the aural cues present in the input audio waveform to separate the individual components. Conventional audio signal processing techniques rely on strong assumptions such as the number of sources in the audio waveforms to compute non-negative matrix factorization~\cite{wang2005musical,canadas2014percussive,fitzgerald2005shifted} of audio spectrograms. Deep learning methods  commonly adopt the self-supervised `mix-and-separate' strategy where multiple audio sources are combined into a synthetic mixture and then predict a spectrogram mask to retrieve queried audio components \cite{zhao2018sound, gao2019co, chatterjee2021visual}.

\noindent\textbf{Multimodal source separation.} Recent work in audio-text \cite{audiocaps,kilgour2022text} and audio-visual \cite{zhao2018sound, gao2019co, zhao2019sound, chatterjee2021visual,chowdhury2021listen} separation also use the `mix-and-separate' strategy to train a decoder to predict a spectrogram mask based on natural language and video queries, respectively. In the case of the former, state-of-the-art approaches can either accept a single object label \cite{zhao2018sound,gao2019co,chatterjee2021visual} or free-form natural language queries \cite{audiocaps,kilgour2022text}. Existing audio-visual source separation approaches often rely on training object detectors with strong supervision from bounding box annotations to localize objects of interest before learning to perform source separation at the object-level. In contrast, our proposed approach does not rely on pretrained object detectors or object labels in the training videos. The Sound of Pixels (SOP) model is the most similar in spirit to our proposed approach since it does not rely on object detectors. Instead, it learns to perform sound separation based on a video-level representation during training. Our \modelabb approach is also similar to the Voiceformer \cite{rahimi2022reading} approach which aims to separate speech from multiple speakers using a combination of audio, visual and language inputs. In contrast to our approach, their approach requires ground-truth text transcripts for training. Finally, the task is also similar in nature to that of sound localization in videos~\cite{feng2022sslnet,hu2022mix}.

\noindent\textbf{Multimodal representation learning.} One key challenge is grounding information from the text and/or audio modalities in the video frames \cite{tan2021look,gao2019co,chatterjee2021visual,anne2017localizing}. Since annotating video datasets is a costly process, prior work has investigated self-supervised pretraining methods on unlabeled videos~\cite{han2020memory, tong2022videomae}. This often involves utilizing complementary information from the audio, text and visual modalities~\cite{zellers2022merlot} to learn robust representations that can generalize to downstream tasks such as action recognition \cite{rouditchenko2020avlnet,piergiovanni2020evolving,akbari2021vatt} and text-to-video retrieval \cite{anne2017localizing, miech2019howto100m}. These approaches often use large-scale pretraining datasets such as Kinetics-400 / 600~\cite{carreira2018short} and HowTo100M~\cite{miech2019howto100m} before finetuning linear probes for target tasks and datasets. Recent work has also focused on prompting multimodal foundation models such as CLIP~\cite{radford2021learning} and ALIGN~\cite{jia2021scaling} to adapt them to new tasks ranging from open-vocabulary object detection \cite{gu2021open} to domain adaptation \cite{gal2022stylegan} without modifying their parameters.


\section{Video-Audio Separation through Text}
\begin{figure}[t]
    \centering
    \includegraphics[width=0.47\textwidth]{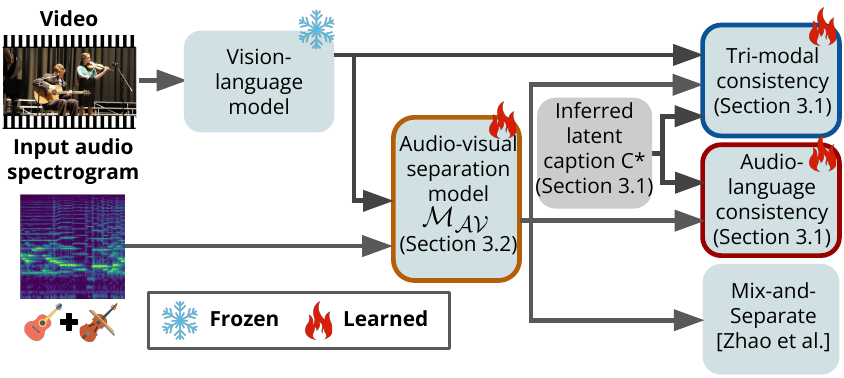}
    \caption{\textbf{Training overview of our proposed \modelabb approach. 
    } We introduce tri-modal (Section~\ref{sec:alignment_losses}) and audio-language (Section~\ref{sec:alignment_losses}) consistency objectives to learn the latent alignment between the audio, video and language modalities. Furthermore, we adopt a prior mask prediction loss~\cite{zhao2018sound} (Section~\ref{sec:model_arch}) to guide the training of our mask prediction decoder. }
    \label{fig:model_overview}
    \vspace{-10pt}
\end{figure}

\begin{figure*}[t]
     \centering
     \begin{subfigure}[b]{0.43\textwidth}
         \raisebox{5mm}{
         \centering
         \includegraphics[width=\textwidth]{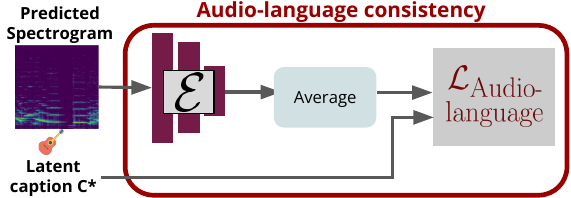}}
         \caption{Audio-language consistency loss.}
         \label{fig:cons_loss}
     \end{subfigure}
     \hspace{3mm}
     \begin{subfigure}[b]{0.51\textwidth}
         \centering
         \includegraphics[width=\textwidth]{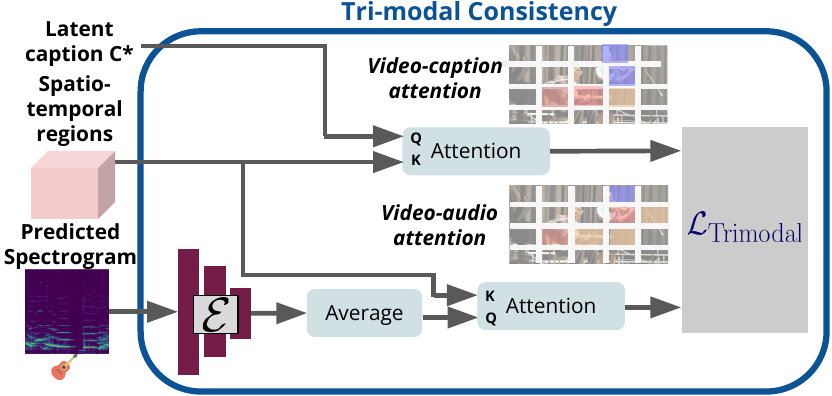}
         \caption{Tri-modal consistency loss.}
         \label{fig:kl_loss}
     \end{subfigure}
     \vspace{-5pt}
        \caption{\textbf{Audio-language and tri-modal consistency alignment objectives.} (a) The audio-language objective aims to maximize the semantic similarity between the predicted audio and its latent caption. (b) In the tri-modal consistency objective, we compute normalized probability distributions over the spatiotemporal regions based on cosine similarity scores (colored frame patches) with respect to the predicted audio and latent captions. $K$ and $Q$ denote the key and query terminology that is used in self-attention. Intuitively, this objective encourages agreement between where the latent caption and predicted audio is occuring in the video.}
        \label{fig:alignment_objs}
        \vspace{-10pt}
\end{figure*}

Given a set of unlabeled videos and their audio tracks, we strive to learn a self-supervised model that separates an audio source given a natural language query and, optionally, an accompanying video. This goal requires a model to learn a strong alignment between the audio, vision, and language modalities. To this end, we propose a novel \modelname (\modelabb) approach that learns an effective representation for audio-visual source separation, as well as a strong alignment between the audio and language modalities for audio-text source separation (see Figure~\ref{fig:model_overview} for an overview). To learn this representation, we exploit the strong joint visual-semantic alignment embodied in pretrained vision-language foundation models by encouraging our model to learn a direct projection function for audio inputs into their joint embedding space (Section~\ref{sec:alignment_losses}). Our key insight is to use videos as an intermediary modality to learn a \emph{transitive} alignment between the audio and language modalities. Since the visual and language modalities are already aligned, learning a strong correspondence between audio and vision should also provide an alignment between audio and language through videos via transitivity. 

Despite the intuitiveness of this idea, our experiments will show that using the visual representations of existing audio-visual foundation models does not help them learn the above-mentioned transitive alignment. To address this limitation, we introduce our novel tri-modal and audio-language consistency  alignment objectives (Figure~\ref{fig:model_overview}). The tri-modal objective encourages the model to maximize the semantic consistency between all three modalities by using videos as the intermediary modality and the audio-language objective helps to further improve the transitive alignment between audio and natural language. We leverage vision-and-language foundation models to infer latent captions that correspond to the audio components without text annotations, thus providing pseudo-language supervision (Section~\ref{sec:alignment_losses}). In addition, we adopt the ``mix-and-separate'' strategy \cite{zhao2018sound} to train our audio separation model, where the audio from multiple videos are mixed and then separated using only the video's visual information. For an input audio waveform $A$, we apply a Short-Time Fourier Transform (STFT) to obtain its magnitude spectrogram $A^S$ and phase $A^{\text{phase}}$.
Our learning objective only makes use of $A^S$, which encodes how the magnitude of the audio frequencies changes over time.
We reconstruct the predicted waveform by applying the inverse STFT on the predicted spectrogram using the original phase $A^{\text{phase}}$. We describe our \modelabb audio separation model $\mathcal{M}$ in Section~\ref{sec:model_arch}.

\subsection{Alignment between audio, language and video}\label{sec:alignment_losses}
A standard method to learn a strong alignment between the audio, language and vision modalities is to maximize the pairwise similarities of their input representations using a contrastive learning formulation \cite{akbari2021vatt,miech2019howto100m}. However, learning this alignment necessitates the presence of labels that indicate ground-truth correspondences between these modalities. In our self-supervised setting, the unlabeled videos only provide noisy labels of audio-visual pairings but do not contain text labels. To circumvent this limitation, we propose to extract \emph{latent captions}, which are \emph{pseudo-words} that encode the visual concepts in videos with language semantics. Prior work \cite{cohen2022my,gal2022image} has demonstrated that the rich visual-semantic alignment in vision-language foundation models can be adapted to extract latent word representations that convey the semantics of their conditioning images. Inspired by this insight, we introduce the novel idea of using latent captions to identify sound-emitting object candidates in unlabeled videos for training, thereby allowing us to train without prior knowledge of existing objects in the videos.

\noindent\textbf{Latent captions.} We perform ``textual inversion" \cite{gal2022textual,gal2022image,cohen2022my} where we adapt the CLIP language encoder to express the visual concepts of a video in the form of an encoded learnable parameter.  Instead of introducing new concepts into the vocabulary of these models which require their parameters to be updated, we search for their latent language representations through a visual reconstruction objective. This objective is based on the key insight that the final learnt latent captions should be expressive enough to convey the visual semantics of the video. While it is also possible to use the visual frame representations as latent captions, we reason that learning them as outputs of the language encoder will help a model generalize better to language queries during inference. For a given video $V$, we encode its center frame $V_{center}$ as: $f^V_{center} = g^V(V_{center})$, where $g^V$ is the embedding function of the CLIP visual encoder. Then, we replace the token embeddings used for language queries with a learnable parameter $p$ and encode it as: $g^L(p)$, where $g^L$ is the embedding function of the CLIP language  encoder. We optimize the weights of $p$ by maximizing the cosine similarity between $f^V_{center}$ and $g^L(p)$:
\begin{equation} \label{eq:latent_caption_opt}
    p^* \in \argmax_{p} sim\left(f^V_{center}, g^L(p)\right)
    \vspace{-5pt}
\end{equation}
where $sim(x,y)=x^T y/(\|x\| \|y\|)$ and $||.||$ denotes the $L_2$ norm operator. Let $C^*=g^L(p^*)$ be the latent caption. We illustrate this operation in the supplemental.
In theory, $C^*$ is analogous to a representation of a description of the frame and may replace ground-truth annotations effectively.

Our audio-language and tri-modal consistency alignment objectives are based on our reasoning that well-separated audio sources should be semantically consistent with the visual concepts that guide their separation as well as their text labels. Given an input video $V$ of $T$ frames $V = \{V_1, \cdot \cdot \cdot V_T \}$ and an audio spectrogram $A^S \in \mathbb{R}^{F \text{x} N}$ with $F$ frequency bins and $N$ STFT time frames, our model predicts a separated audio spectrogram $\hat{A}^S \in \mathbb{R}^{F \text{x} N}$. We extract a latent representation for the predicted spectrogram: $\hat{f}^A = \mathcal{M}_\theta(\hat{A}^S) \in \mathbb{R}^{D}$ where $\mathcal{M}$ denotes our audio separation model. For each video, we use its encoded predicted spectrogram and latent caption to provide pseudo-language supervision in our alignment objectives detailed below.

\smallskip
\noindent\textbf{Audio-language consistency loss.} To encourage our audio encoder to learn audio representations that are effective for separating sound components when conditioned on either text or video queries, we aim to maximize the pairwise similarities between all three modalities. The key insight is that the audio sources that are well-separated by the visual concepts in a video should have a strong semantic relevance to its latent caption which express the same concepts in natural language (Figure~\ref{fig:cons_loss}). Theoretically, this is similar to the self-supervised multimodal learning objective of maximizing the similarity between the representations of an image and its corresponding caption as well as the dissimilarity of non-corresponding captions. In lieu of ground-truth object labels, we can maximize the alignment between the predicted audio representations and the latent captions over the entire vocabulary of captions $\mathcal{X}$:
\begin{equation}
    \mathcal{L}_{\text{Audio-language}} =
-\log \left(\frac{\exp(\hat{f}^A \cdot C^*/\tau)}{\sum_{x\in\mathcal{X}} \exp(\hat{f}^A  \cdot x / \tau)} \right)
\end{equation}
where $\tau$ is the temperature. 

However, the problem of false negatives has been well-documented in image-text datasets, where captions for some images may be relevant to other images but are treated erroneously as negatives. Since we are training on unlabeled videos, we account for the high likelihood that some videos may contain similar sounding objects by using a lower weight for this objective. Intuitively, this weighting helps to alleviate the effect of false negatives and prevent it from introducing a lot of noise during training.

\noindent\textbf{Tri-modal consistency loss.} While the audio-language consistency objective facilitates improving the alignment between audio sources and their corresponding latent captions, it does not provide a straightforward solution to disregard false negatives in its contrastive formulation. To address this issue, we further introduce a softer tri-modal alignment constraint which exploits the implicit localization capability of vision-language foundation models for supervision without requiring any negative samples. Specifically, we propose to use the intermediary visual modality to encourage well-separated audio components to be grounded correctly in the relevant spatiotemporal regions of a video. Prior work \cite{zhou2021denseclip,rao2022denseclip} has demonstrated that the CLIP model can be engineered to extract a segmentation map based on natural language queries. Inspired by this finding, we use the latent captions to provide pseudo-target attention maps.

Our intuition is that enforcing a soft constraint on the predicted spectrograms such that they can be mapped to similar relevant regions as determined by the latent captions will encourage the model to implicitly learn the transitive alignment between the audio and language modalities.
Let $P_{Att}(A, b) = \sigma\left(A \cdot b\right)$ to
be the attention operation where $\sigma(z)_i = z_i/\sum_j z_j$ is the softmax function. For a given video $V$, we extract a set of spatiotemporal region representations by encoding each frame separately. Specifically, we encode the $t$-th frame as: $f^V_{t} = g_\theta^V(V_t)$, where $f^V_t \in \mathbb{R}^{HW \text{x} D}$, where $H$ and $W$ are the downsampled height and width of the input video. Then, we compute a probability distribution of similarity scores over the set of spatial region features with respect to its latent caption $C^*$ for the $t$-th frame: $P_{VC}^t = P_{Att}(f^V_t, C^*) \in \mathbb{R}^{HW\times 1}$.
Next, we compute a similar distribution over the region features of the $t$-th frame with respect to the encoded audio representation of the masked spectrogram: $P_{VA}^t = P_{Att}(f^V_t, \hat{f}^A) \in \mathbb{R}^{HW\times 1}$. We encourage the predicted audio representation to be similar to that of the latent caption by minimizing the Kullback-Leibler (KL) divergence between both attention distributions over the set of all time steps in a video $\mathcal{T}$:
\begin{equation}
    \mathcal{L}_{\text{Trimodal}} = \mathbb{E}_{t\sim\mathcal{T}}\left[\sum_{x=1}^{HW} P_{VC}^t(x)\log\frac{P_{VC}^t(x)}{P_{VA}^t(x)}\right],
\end{equation}
where $P_{VA}^t(x)$ denotes the audio-video attention distribution over the $t$-frame of the $x$-th spatial region.
\begin{figure}[t]
    \centering
    \includegraphics[width=0.5\textwidth]{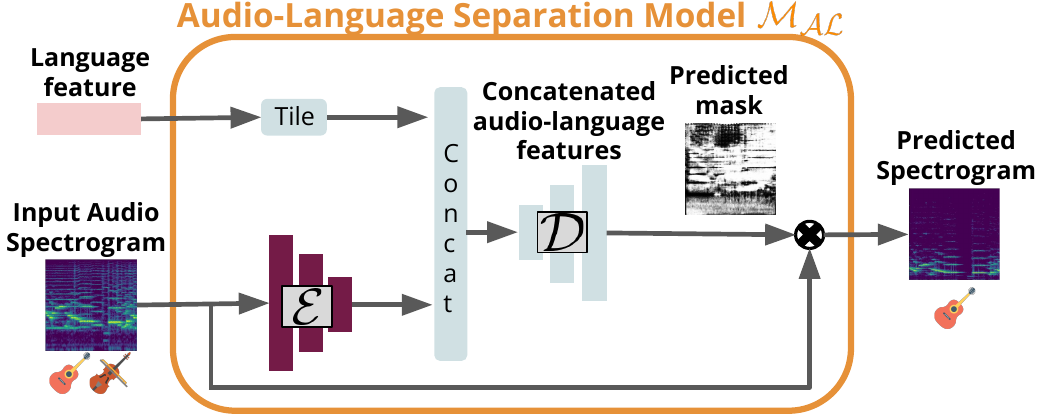}
    \caption{\textbf{Spectrogram mask prediction.} During inference, our model is able to separate audio sources using video or language queries despite training without ground-truth text annotations. We note that we only use unlabeled videos during training. We illustrate the audio-visual separation mode $\mathcal{M}_{AV}$ in the supplemental.}
    \label{fig:forward_pass}
    \vspace{-5pt}
\end{figure}

\begin{table*}[t]
\centering
\begin{tabular}{|l|c|c|ccc|ccc|}
\hline
& Object   & \# Region &   &  Solos  &   &    &  Duets  &  \\
Method   &  detectors & proposals & SDR $\uparrow$  & SIR $\uparrow$ & SAR $\uparrow$ &  SDR $\uparrow$ & SIR $\uparrow$ & SAR $\uparrow$\\
\hline
Co-Separation \cite{gao2019co} & Yes & 2 & 7.38 & 13.70 & 10.82 & 7.42 & 13.81 & 10.60 \\
AVSGS\tablefootnote{\text{Compute-efficient.}} \cite{chatterjee2021visual} & Yes & 42 & \textbf{9.04} & \textbf{14.45} & \textbf{12.24} & \textbf{10.25} & \textbf{15.60} & \textbf{12.82} \\
\hline
Sound-of-Pixels \cite{zhao2018sound} & No & None & 7.30 & 11.90 & 11.90 & 6.05 & 9.81 & 10.61\\
AV-Mix-and-Separate~\cite{gao2019co} & No & None & 3.16 & 6.74 & 8.89 & 3.23 & 7.01 & 9.14\\
NMF-MFCC \cite{spiertz2009source} & No & None & 0.92 & 5.68 & 6.84 & 0.92 & 5.68 & 6.84\\
\modelabb (Ours) & No & None & \textbf{7.98} & \textbf{13.92} & \textbf{12.35} & \textbf{8.08} & \textbf{13.97} & \textbf{11.33} \\
\hline 
\end{tabular}
\caption{\textbf{Audio-visual source separation on MUSIC.} We report results on videos that contain one instrument (solos) and two instruments (duets). Despite not training with object bounding boxes, \modelabb performs competitvely with state-of-the-art detection-based approaches.}
\label{tab:music_eval}
\vspace{-10pt}
\end{table*} 

\subsection{Audio Separation Model}\label{sec:model_arch}
We introduce two audio separation models $\mathcal{M}_{AL}$ and $\mathcal{M}_{AV}$ which share the same audio model parameters for separating audio sources. $\mathcal{M}_{AL}$ (Figure~\ref{fig:forward_pass}) infers a spectrogram mask based on a language query while $\mathcal{M}_{AV}$ (shown in supplemental) conditions its prediction on the input video. 
We adopt the U-Net \cite{ronneberger2015u} model 
 as the audio component in $\mathcal{M}_{AL}$ and $\mathcal{M}_{AV}$, which has been demonstrated by prior work \cite{zhao2018sound, gao2019co} to be effective at audio source separation, and the CLIP visual and language encoders to exploit its learnt visual-semantic alignment. The U-Net model comprises an encoder $\mathcal{E}$ and a decoder $\mathcal{D}$. Given an input audio spectrogram $A^S$, the encoder $\mathcal{E}$ applies a series of downsampling convolutional layers to output a set of bottleneck features: $f^A = \mathcal{E}(A^S), f^A \in \mathbb{R}^{H^A \times W^A \times D}$, where $H^A$ and $W^A$ denote the height and width of the downsampled spectrogram, respectively. 

Given a natural language query $L \in \mathbb{R}^{N_w}$ where $N_w$ is the number of words, we extract its representation $f^L \in \mathbb{R}^{1 \text{x} D}$ using the text encoder: $f^L = g^L(L)$. We stress that language queries with ground-truth object labels are only used during inference. Then, we tile the language representation by the factor $H^A W^A$ and concatenate with the audio bottleneck representations along the channel dimension: $f^{AL} = concat(f^A, tile(f^L))$, where $f^{AL}$ has the dimensions $\mathbb{R}^{H^A \times W^A \times 2D}$. We pass the concatenated representations into the decoder $\mathcal{D}$ consisting of a series of upsampling convolutional layers to generate a real-valued ratio mask: $\hat{M} = \mathcal{D}(f^{AL}) \in \mathbb{R}^{F \times N}$. To predict the separated audio source, each element of the mask is multiplied with the corresponding location in the input spectrogram: $\hat{A}^S = \hat{M} \odot A^S$, where $\odot$ denotes the Hadamard product.  

\smallskip
\noindent\textbf{Video and Multiple Instance Learning for mask prediction.} Since language annotations are not used during training, we train only with unlabeled videos. During training, as well as inference on audio-visual separation, we use the CLIP visual encoder to extract a set of language-grounded spatiotemporal region representations $f^V \in \mathbb{R}^{T \times H \times W \times D}$. Similarly, we can perform the same mask inference operation using videos by using the representation of each spatiotemporal region representation interchangeably in place of the language representation. This interchangeability is due to the fact that the CLIP representations project the video to the joint vision-language embedding space.

Prior work \cite{gao2019co} has demonstrated the advantages of performing audio-visual separation at the object level for videos with multiple objects. In the absence of object detectors and labels, we propose an MIL formulation, where all spatiotemporal regions in the input video are treated as positive candidates. Specifically, given a spatiotemporal grid of region representations extracted from the visual encoder as defined in this section,  we predict a spectrogram mask $\hat{M}_{j}$ for the $j$-th region. 
Finally, we aggregate them over all regions to obtain a spectrogram mask for the entire video: 
\begin{equation}
    \hat{M} = \frac{1}{T}\sum\nolimits_{j=1}^{THW}\hat{M}_{j}.
\end{equation}
Intuitively, this formulation encourages the model to learn to identify the salient regions with sounding objects for audio separation. We obtain the predicted mask for each frame by computing the sum of the predicted masks over all spatial regions and average the predicted masks over the temporal dimension since we reason that a sounding object should have similar prediction masks across frames. We adopt the self-supervised `mix-and-separate' learning objective \cite{zhao2018sound,gao2019co} to train our model for predicting spectrogram masks. In this formulation, audio from two or more videos are mixed and the goal is to use the visual information of a video to separate its audio component. This setting allows us to generate pseudo-target masks for training supervision. We compute $\mathcal{L}_{mask}$ as the L1 loss between the pseudo-target and predicted ratio masks \cite{zhao2018sound}. Please refer to the supplemental for details of our spatiotemporal region representation, audio-visual separation model and $\mathcal{L}_{mask}$. The final objective function is computed as:
\begin{equation}
    \mathcal{L} = \lambda_1 \mathcal{L}_{mask} + \lambda_2 \mathcal{L}_{\text{Trimodal}} + \lambda_3 \mathcal{L}_{\text{Audio-language}}
\end{equation}
where $\lambda_1$, $\lambda_2$ and $\lambda_3$ denote the weights for the mask, tri-modal and audio-language consistency losses, respectively.

\section{Experiments}\label{sec:experiments}

We evaluate the effectiveness of our proposed \modelabb approach on audio-text and audio-visual source separation in videos. To perform fair comparison to prior work, we conduct experiments on the SOLOS \cite{montesinos2020solos}, MUSIC \cite{zhao2018sound} and AudioSet \cite{gemmeke2017audio} datasets. We refer readers to the supplemental for additional implementation and dataset details. Given a language query containing a prompt template and an object, the goal of \textbf{audio-text source separation} is to extract its corresponding sound component from an audio input, which is often comprised of a mixture of sounds emitted by various objects. Similarly, in \textbf{audio-visual source separation}, a model has to leverage the visual information in the input video to extract its corresponding sound component and align it with the relevant video pixels.

\noindent\textbf{Evaluation metrics.} We adopt the Signal-to-Distortion Ratio (SDR), Signal-to-Interference Ratio (SIR) and Signal-to-Artifact Ratio (SAR) \cite{raffel2014mir_eval,rix2001perceptual}. SDR computes the difference between the SDR of the predicted and mixed audio with respect to the ground-truth waveforms, which measures the sound quality of the separated audios. SIR measures the amount of contamination from other audio sources in the predicted audio. Finally, SAR measures the amount of artifacts in the predicted audio.



\subsection{Quantitative results}\label{sec:results}

We compare our approach to the following baselines. Similar to our approach, \textbf{Sound of Pixels (SOP)} \cite{zhao2018sound} does not require object detectors but learns to perform video-level separation. We also implement the \textbf{AV-Mix-and-separate} baseline from \cite{gao2019co} with the same audio-visual separation model as ours but is trained to perform video-level separation and without our alignment objectives. The off-the-shelf \textbf{NMF-MFCC} \cite{spiertz2009source} baseline performs audio separation using non-negative matrix factorization (NMF) and Mel frequency cepstrum coefficients (MFCC). Next, we also compare to \textbf{Co-Separation} \cite{gao2019co} that learns to perform audio separation at the object-level by relying on object detectors. Last but not least, we include \textbf{AVSGS} \cite{chatterjee2021visual} that also relies on a high number of object proposals to construct a spatiotemporal scene graph to reason about context.

\noindent\textbf{Audio-visual source separation on MUSIC.} We report the results of our evaluation under two settings in Table~\ref{tab:music_eval}. In the first setting, each video contains only a single instrument while the `duet' setting includes videos that contain two different instruments. To begin, our \modelabb approach outperforms SOP by a large margin despite not relying on object detectors too. As shown in Table~\ref{tab:sop_visual_comparison}, we show that this performance gain is not due to just replacing ImageNet-pretrained Resnet18 \cite{he2016deep} visual representations with those of the CLIP model since it leads to a large drop in performance. We hypothesize that SOP does not learn a direct projection of the audio inputs used in source separation into the CLIP embedding space. In contrast, \modelabb facilitates a better adaptation of the learnt audio representations to the CLIP embedding. We also found it important to modify CLIP's self-attention (see supp.). We observe that \modelabb outperforms AV-Mix-and-Separate by significantly, particularly under the duet setting. This suggests that performing region-level audio separation provides more capacity for the model to reason about multiple sounding objects in videos.


More significantly, we observe that our approach outperforms the Co-Separation approach \cite{gao2019co}, which relies on object labels in the videos. Since Co-Separation uses a high confidence region proposal for each instrument to localize relevant regions, our improvements over it suggest that our latent captions are able to express multiple visual concepts that are present in the video. Last but not least, our approach is also comparable to AVSGS even without scene graph reasoning modules. The latter constructs a spatiotemporal visual scene graph over a large number of region proposals to reason about context between detected objects. We note that their audio-visual scene graph component can be combined with our \modelabb approach to possibly improve performance but is beyond the scope of this work.

\begin{table}[t]
\centering
\begin{tabular}{|l|c|ccc|}
\hline
& \small{Object}   &    &    &   \\
\small{Method}   &  \small{detectors}  & \small{SDR} $\uparrow$ & \small{SIR} $\uparrow$ & \small{SAR} $\uparrow$ \\
\hline
\small{Co-Separation \cite{gao2019co}} & Yes & 7.11 & 12.09 & 10.05 \\
\small{AVSGS \cite{chatterjee2021visual}} & Yes & \textbf{9.20} & \textbf{14.05} & \textbf{12.17} \\
\hline
\small{Sound-of-Pixels \cite{zhao2018sound}} & No & 6.28 & 10.84 & 10.13 \\
\small{AV-Mix-and-Sep~\cite{gao2019co}} & No & 2.94 & 5.81 & 8.33 \\
\small{NMF-MFCC} \cite{spiertz2009source} & No & 0.68 & 4.75 & 5.12 \\
\small{\modelabb (Ours)} & No & \textbf{8.58} & \textbf{14.16} & \textbf{12.35} \\
\hline 
\end{tabular}
\caption{\textbf{Audio-visual source separation results on the SOLOS dataset.} Similar to the results in Table~\ref{tab:music_eval}, we see that \modelabb significantly closes the gap with object detection-based approaches.}
\label{tab:solos_eval}
\vspace{-5pt}
\end{table}

\begin{table}[h]
\centering
\begin{tabular}{|l|c|ccc|}
\hline
& \small{Object} &    &    &   \\
\small{Method}   &  \small{detectors}  & \small{SDR} $\uparrow$  & \small{SIR} $\uparrow$  & \small{SAR} $\uparrow$\\
\hline
\small{Co-Separation \cite{gao2019co}} & Yes & 4.26 & 7.07 & 13.00 \\
\small{AVSGS \cite{chatterjee2021visual}} & Yes & \textbf{5.28} & \textbf{8.27} & \textbf{13.04} \\
\hline
\small{Sound-of-Pixels \cite{zhao2018sound}} & No & 1.66 & 3.58 & 11.50 \\
\small{AV-Mix-and-Sep~\cite{gao2019co}} & No & 1.68 & 3.30 & 12.20 \\
\small{NMF-MFCC} \cite{spiertz2009source} & No & 0.25 & 4.19 & 5.78 \\
\small{\modelabb (Ours)} & No & \textbf{4.15} & \textbf{7.62} & \textbf{13.20} \\
\hline 
\end{tabular}
\caption{\textbf{Audio-visual source separation results on AudioSet.} \modelabb generalizes much better to the noisier AudioSet dataset than Sound-of-Pixels which also does not use object detections.}
\label{tab:audioset_eval}
\end{table} 

\begin{table}
\begin{center}
\begin{tabular}{|l|c|ccc|}
\hline
Model & Visual encoder & SDR $\uparrow$ & SIR  $\uparrow$ & SAR $\uparrow$ \\
\hline 
SOP \cite{zhao2018sound} & ImageNet  & 6.28 & 10.84  & 10.13 \\
SOP \cite{zhao2018sound} & CLIP & 4.42 & 8.36 & 8.21 \\
Ours & CLIP & \textbf{8.58} & \textbf{14.16} & \textbf{12.35} \\
\hline 
\end{tabular}
\caption{\textbf{Using CLIP visual features in SOP model on the SOLOS dataset.} Using CLIP visual features naively in existing methods results in a drop in performance.}
\label{tab:sop_visual_comparison}
\end{center}
\vspace{-20pt}
\end{table}

\noindent\textbf{Audio-visual source separation on SOLOS.} While the videos in the SOLOS dataset generally have less background noise than those of the MUSIC dataset, we see in Table~\ref{tab:solos_eval} that AV-Mix-and-Separate and NMF-MFCC are still unable to generalize to the cleaner audio signals.  Similar to the reported performance on the MUSIC dataset, we also observe that \modelabb is comparable to the strongly supervised AVSGS approach on the SOLOS dataset. One possible reason behind the lower performance of AVSGS on videos with single instrument is that it is less critical to reason about context for videos with a single sounding object. Our better performance as compared to Co-Separation suggests that learning to perform audio separation at the coarse video region level under an MIL formulation can serve as an effective replacement for training object detectors. These results also indicate the great promise of latent captions in replacing ground-truth object annotations for supervision.

\noindent\textbf{Audio-visual source separation on AudioSet.} Finally, we report the results of our evaluation on the AudioSet dataset in Table~\ref{tab:audioset_eval}. AudioSet has been documented by prior work \cite{gao2019co} to be much noisier than the videos in the other two datasets, which explains the lower performance across all approaches. Unlike the SOP model, we observe that our training approach is more robust to noise in the dataset where the sounding object may not always be visible.

\begin{table}
\centering
\begin{tabular}{|c|c|ccc|}
\hline
$\mathcal{L}_{\text{Audio-language}}$ & $\mathcal{L}_{\text{Trimodal}}$ & SDR $\uparrow$ & SIR $\uparrow$  & SAR  $\uparrow$\\
\hline \xmark & \xmark & 5.47 & 10.55 & 10.95 \\
\cmark & \xmark & 8.08 & 13.74 & 12.18 \\
\xmark & \cmark & 8.10 & 13.84 & 11.79 \\
\cmark & \cmark & \textbf{8.58} & \textbf{14.16} & \textbf{12.35} \\
\hline 
\end{tabular}
\caption{\textbf{Ablation of our audio-language and tri-modal consistency alignment objectives on SOLOs.} The alignment objectives help to improve audio-visual separation performance.}
\label{solos_sop_loss_weights_ablation}
\end{table}

\begin{table}
\centering
\begin{tabular}{|c|ccc|}
\hline
\# tokens  & SDR $\uparrow$ & SIR $\uparrow$ & SAR $\uparrow$ \\
\hline 
1  & 7.31 & 11.34 & 11.71\\
2  & 7.67 & 13.07 & 11.45\\
3  & \textbf{8.58} & \textbf{14.16} & \textbf{12.35}\\
4 & 8.02 & 13.39 & 11.53\\
\hline 
\end{tabular}
\caption{\textbf{Ablation over number of learnable tokens on SOLOS.} Using more learnable tokens generally improves performance.
}
\label{solos_num_learnable_tokens_ablation}
\vspace{-15pt}
\end{table}

\begin{table*}[t]
\centering
\begin{tabular}{|c|c|ccc|ccc|ccc|}
\hline
Query & Alignment  &    & SOLOS &   &    & MUSIC & &    & Audioset & \\
Modality   &  objectives  & SDR $\uparrow$ & SIR $\uparrow$ & SAR $\uparrow$ & SDR $\uparrow$ & SIR $\uparrow$ & SAR $\uparrow$ & SDR $\uparrow$ & SIR $\uparrow$ & SAR $\uparrow$ \\
\hline
Language & No & -3.05 & 2.79 & 3.77 & -3.67 & 2.51 & 3.41 & -5.02 & 1.98 & 14.94\\
Language & Yes & \textbf{6.92} & \textbf{12.07} & \textbf{10.41} & \textbf{6.45} & \textbf{11.18} & \textbf{10.77} & \textbf{2.36} & \textbf{4.71} & \textbf{10.28} \\
\hline 
\end{tabular}
\caption{\textbf{Audio-text separation with natural language queries.} We evaluate our model, that is trained only on unlabeled videos, on the task of audio-text separation. Note that we do not compare to other audio-visual separation baselines since there is no straightforward way to adapt them for language queries. We compare our full \modelabb model to a variant that is only trained with the mask prediction loss.}
\label{tab:language_queries_eval}
\vspace{-10pt}
\end{table*}
\vspace{-10pt}
\subsubsection{Model Analysis}
\noindent\textbf{Ablation of $\mathcal{L}_{\text{Trimodal}}$ and $\mathcal{L}_{\text{Audio-language}}$. } Table~\ref{solos_sop_loss_weights_ablation} provides an ablation over our audio-language and tri-modal consistency losses. Despite being trained on a larger and more diverse set of visual concepts, we find the CLIP visual representations alone do not encourage the audio encoder to learn a strong alignment between the audio and vision modalities (row 1). We observe the importance of our proposed alignment objectives where adding one or the other leads to a significant improvement in audio separation performance. Due to multiple training videos containing similar instruments, it is likely that we are treating some latent captions as false negative samples for each video in the audio-language consistency objective. However, we see that the consistency objective is still beneficial towards audio separation.

\noindent\textbf{Number of learnable token embeddings in latent captions.} We report the results of our ablation over the number of learnable token parameters used in the extraction of latent captions in Table~\ref{solos_num_learnable_tokens_ablation}. We observe that increasing the number of learnable tokens generally helps to improve the performance, although using 4 tokens hurts performance slightly. We hypothesize that a higher number of learnable tokens provide more capacity for the CLIP language model to express multiple visual concepts that are present in the videos. This finding suggests that using more tokens may be beneficial for videos with more complex visual scenes. 


\subsection{Audio-language source separation}
Finally, we evaluate the capability of our trained model to separate audio components based on natural language queries in Table~\ref{tab:language_queries_eval}. We construct each query using the template `person playing a \{instrument\}'. We stress that we only train on unlabeled videos without text supervision. To begin, we compare our \modelabb model to a variant that is trained without our alignment objectives. Note that we do not compare to other audio-visual approaches since there is no straightforward way to adapt them for language queries. As evidenced by the significant performance gap between both variants, our alignment losses are integral to learning a strong transitive alignment between the audio and language modalities. This finding suggests that just learning an alignment between the audio and visual modalities does not result in a \emph{transitive} relationship between the audio and natural language modalities. 

Additionally, for effective sound source separation given either video or text inputs, we hypothesize that it is critical to learn an audio encoder E that is well-aligned with CLIP's vision-language embedding. Note that the same shared audio encoder $\mathcal{E}$ encodes both mixed and predicted audio sources. Thus, it can also be used as a general audio encoder and the decoder $\mathcal{D}$ can condition its separation process on either language or video queries. Lastly, we observe that our alignment objectives are still insufficient for closing the performance gap between audio-text and audio-visual separation (Tables~\ref{tab:music_eval}, \ref{tab:solos_eval} and \ref{tab:audioset_eval}) completely. This observation indicates that future work is needed for further improving the latent alignment between all three modalities.

\subsection{Qualitative results}
\begin{figure}[h]
    \centering
\includegraphics[width=0.5\textwidth]{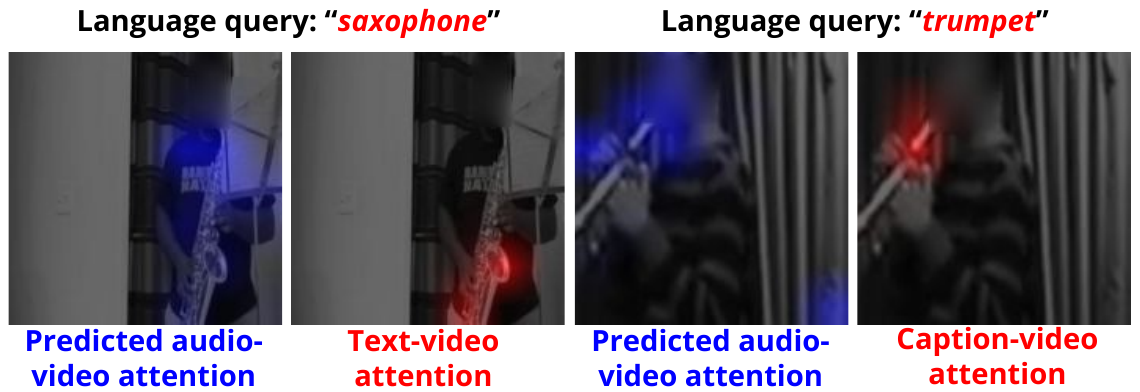}
    \caption{\textbf{Predicted audio attention in videos.}
    Our consistency objectives help the model to localize the relevant regions more accurately without prior knowledge of sounding object locations.
    }
    \label{fig:audio-to-video-attn}
    \vspace{-10pt}
\end{figure}

\noindent\textbf{Predicted audio-to-video attention visualizations.} To evaluate the semantic consistency between the predicted audio sources and the visual cues that guide the separation process, we plot the attention map between the encoded predicted spectrograms and their corresponding video frames (Section~\ref{sec:alignment_losses}) in Figure~\ref{fig:audio-to-video-attn}. We observe that \modelabb helps the model to learn to ground the audio source in the video pixels despite not requiring bounding box supervision.

\begin{figure}[h]
    \centering
\includegraphics[width=0.5\textwidth]{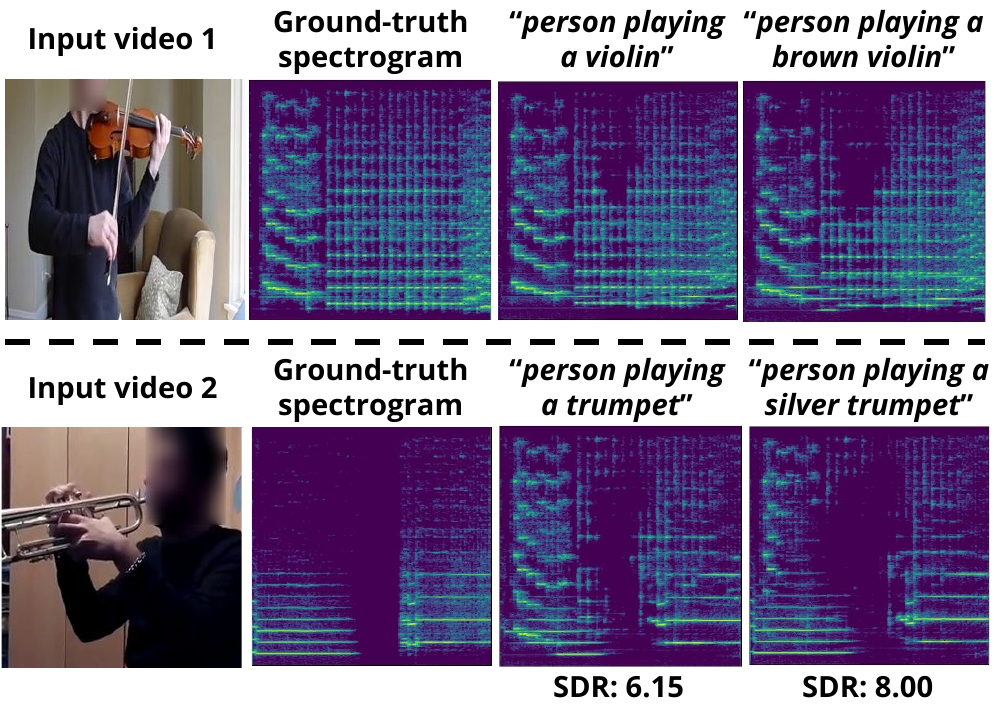}
    \vspace{-20pt}
    \caption{\textbf{Predicted audio spectrograms with language queries.} Interestingly, we see that including visually grounded adjectives in the queries affects the quality of the separated audio sources.
    }
    \label{fig:language_spec_outputs}
    \vspace{-15pt}
\end{figure}

\noindent\textbf{Predicted outputs of language-based audio separation.} Finally, we present an example of how different language queries affect the separation performance for the same audio input in Figure~\ref{fig:language_spec_outputs}. Significantly, despite not relying on text annotations for training supervision, our model can separate audio components based on free-form language queries. This demonstrates the effectiveness of our proposed idea to substitute ground-truth text annotations with \emph{latent captions} during training. Additionally, we also observe that providing more context in the language query (\eg, ``violin" versus ``person playing a violin") can lead to improved separations.

\vspace{-5pt}

\section{Conclusion}
In summary, we introduce a self-supervised approach for learning to separate audio based on a language query, or a video. In the absence of object labels, we propose to extract latent captions to provide pseudo-language supervision. Furthermore, we introduce the novel tri-modal and audio-language alignment objectives that use the latent captions to improve the alignment of the audio, language and video modalities. By reducing the training need for object labels, our work opens up the possibility of large-scale pretraining on unlabeled videos with diverse visual concepts. 

\noindent 
\textbf{Acknowledgements}: This material is based upon work supported, in part, by DARPA under agreement number HR00112020054 and a gift from Adobe Research.

{\small
\bibliographystyle{ieee_fullname}
\bibliography{egbib}
}

\clearpage
\appendix

In this supplemental, we provide the following additional material to the main paper:
\begin{enumerate}
    \item[A] Latent caption extraction details
    \item[B] Extraction of CLIP region representations
    \item[C] Mix-and-separate training strategy and $\mathcal{L}_{\text{mask}}$
    \item[D] Dataset details
        \begin{enumerate}
            \item MUSIC
            \item SOLOS
            \item AudioSet
        \end{enumerate}
    \item[E] Implementation details
    \item[F] Additional ablation experiments
        \begin{enumerate}
            \item weights for $\mathcal{L}_{\text{Audio-language}}$ and $\mathcal{L}_{\text{Tri-modal}}$
            \item shared parameters for audio U-Net encoder $\mathcal{E}$
            \item Bounding box experiment
        \end{enumerate}
    \item[G] Discussion of limitations of \modelabb
    \item[H] Predicted separated audio samples
\end{enumerate}

\section{Latent caption extraction}
\begin{figure}[h]
    \centering
    \includegraphics[width=0.5\textwidth]{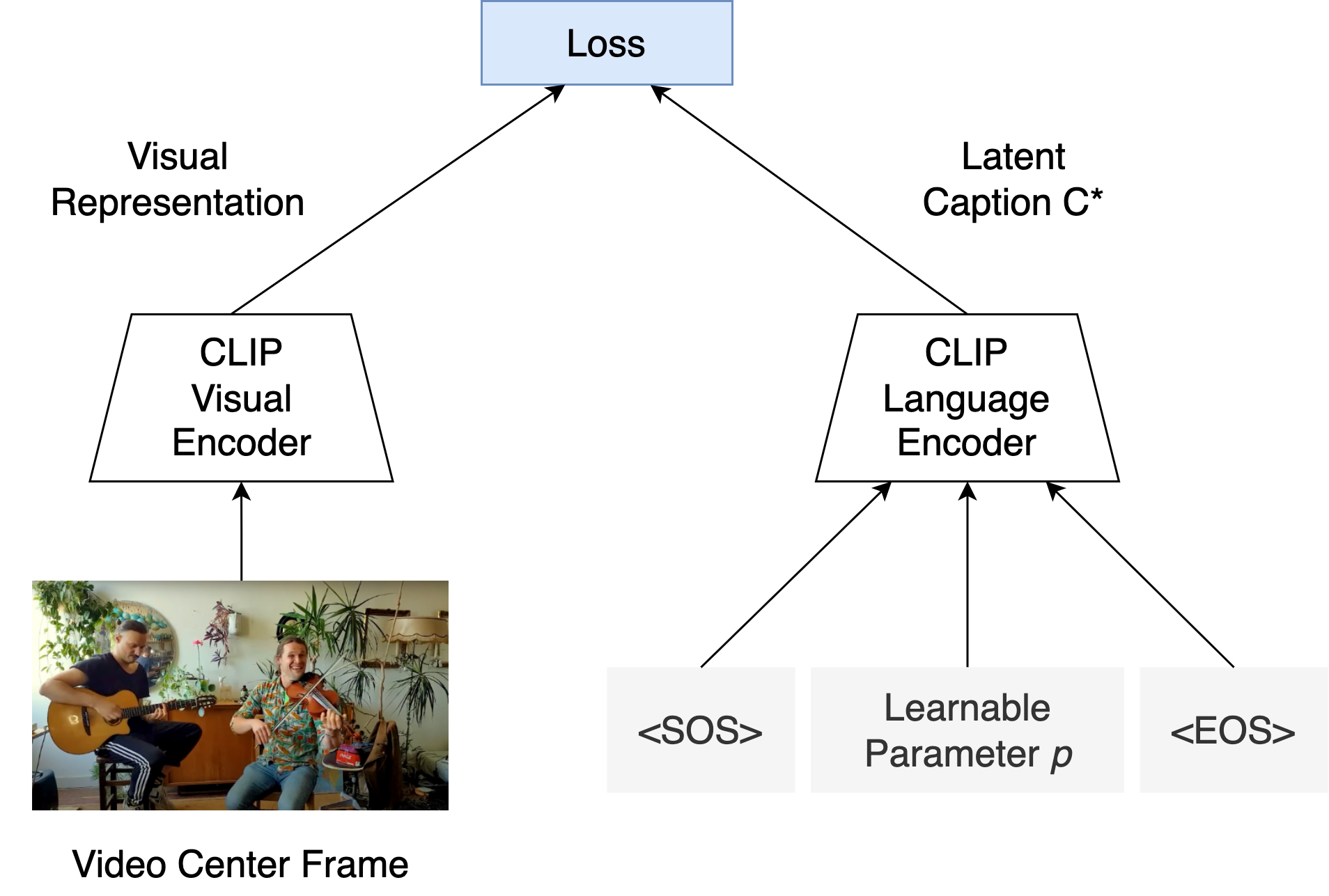}
    \caption{\textbf{Extraction of latent captions for pseudo-supervision.} We formulate the extraction mechanism as an optimization process and learn the weights of the parameter $p$ by maximizing the cosine similarity between the final visual and language representations.}
    \label{fig:latent_extraction}
\end{figure}

We provide an illustration of our latent caption extraction operation (Section \textcolor{red}{3.1}) in Figure~\ref{fig:latent_extraction} and a more detailed description of the entire operation. As mentioned earlier, we extract a latent caption from each unlabeled video to provide pseudo-language supervision. Given a video $V$, we begin by encoding its center frame using the CLIP visual encoder: $f^V_{\text{center}} = g^V(V_{\text{center}})$. Symmetrically, we seek to extract a language representation that corresponds to the encoded center frame semantically, described next. 

The encoding function of the CLIP language transformer encoder $g^L$ provides a mechanism that is amenable to searching for latent captions that already exist in its learnt vocabulary, which allows us to freeze its parameters and leverage its strong visual-semantic alignment with the vision modality. Instead of using the trained token embeddings, we introduce a learnable token parameter $p$ and pass it into the language encoder $g^L$. We adopt the simple objective function of maximizing the cosine similarity between the center frame representation and the output of the language encoder, which allows us to update the weights of $p$ through gradient back-propagation. We formulate the optimization operation mathematically as:
\begin{equation}
    p^* = \argmax_{p} sim\left(f^V_{center}, g^L(p)\right)
\end{equation}
where $sim(x,y)=x^T y/(\|x\| \|y\|)$ and $||.||$ denotes the $L_2$ norm operator. We compute the final latent caption of the video as $C^* = g^L(p^*)$. The latent captions are used in our proposed alignment objectives to provide pseudo-language supervision. The search time for parameter $p$ in Equation \textcolor{red}{1} is about $\sim$148 seconds per video on a RTX 2080 GPU for 5k iterations.

\section{Extraction of spatiotemporal region representations from CLIP in Section \color{red} 3.1} \label{sec:latent_extraction}
We begin by providing an overview of the 2D attention pooling layer in the CLIP Resnet visual encoders. By default, the CLIP visual encoder outputs a global visual representation for each input image. While we use the Resnet variants instead of the transformer-based architectures in CLIP, the former differs from the standard Resnet architecture in two ways. First, the CLIP variant contains three convolutional stems instead of one. Second, and more importantly, the CLIP Resnet variant also replaces the global average pooling (GAP) layer with a 2D self-attention operation, which contains the key, query and value projections. Next, we describe in more detail this self-attention layer and how we modify it for our task.

\noindent\textbf{CLIP 2D attention pooling.} We begin by extracting a set of spatial region representations from an input image $\mathcal{I}$ as: $f^I = g^V(I) \in \mathbb{R}^{HW \times D}$, where $H$, $W$ and $D$ are the downsampled height, width and channel dimensions. Recall that a self-attention operation involves the use of keys, queries, and values. The CLIP model computes an average image representation as the query vector: $\overline{f}^I = \frac{1}{HW} \sum\limits_{j=1}^{HW} f^I_j$, where $f^I_j$ denotes the $j$-th row of $f^I$. Then, it computes a final representation for the entire image as follows:
\begin{equation}
    \begin{aligned}
    K &= \overline{f}^I W_K \in \mathbb{R}^{1 \times D}\\    
    Q &= f^I W_Q \in \mathbb{R}^{HW \times D} \\
    V &= f^I W_V \in \mathbb{R}^{HW \times D}
  \end{aligned}
\end{equation}
where $W_K$, $W_Q$ and $W_V$ are the key, query and value projection matrices, respectively and $W_K$, $W_Q$ and $W_V \in \mathbb{R}^{D\times D}$. Lastly, we compute the final contextualized image representation as:
\begin{equation}
    f^I_{\text{global}} = W_L\left(V^\top \operatorname{softmax}\left(\frac{\left(Q K^\top\right)}{\sqrt{D}}\right)\right)
\end{equation}
where $W_L$ is the final language projection layer that maps the visual representations into the joint visual-semantic embedding space and $W_L \in \mathbb{R}^{D\times D}$ . 

\noindent\textbf{Modified attention operation.} Our Multiple Instance Learning formulation necessitates the presence of region representations in each input frame since we are predicting a spectrogram mask for each region. Additionally, we require these region representations to be well-aligned with the language modality such that a region should have a high similarity with the language query if its visual concept is semantically consistent with that of the query. Consequently, we extract a set of spatiotemporal region representations $f^V_{\text{conv}}$ for our input video $V$ with $T$ frames. We encode the $t$-frame as: $f^V_{t, \text{conv}} = g^V(V_t) \in \mathbb{R}^{HW \times D}$. Finally, we compute the set of language-aligned spatiotemporal region representations by projecting them through the value and language projection layers as follows:
\begin{equation}
    \begin{aligned}
    f^V_{\text{val}} &= W_V f^V_{\text{conv}} \\
    f^V &= W_L f^V_{\text{val}}
    \end{aligned}
\end{equation}
We pass this set of spatiotemporal region representations into our audio separation model $\mathcal{M}$ along with an input audio spectrogram to predict a mask. 

\section{Mix-and-separate training objective in Section \color{red} 3.2}
\begin{figure}[t]
    \centering
    \includegraphics[width=0.45\textwidth]{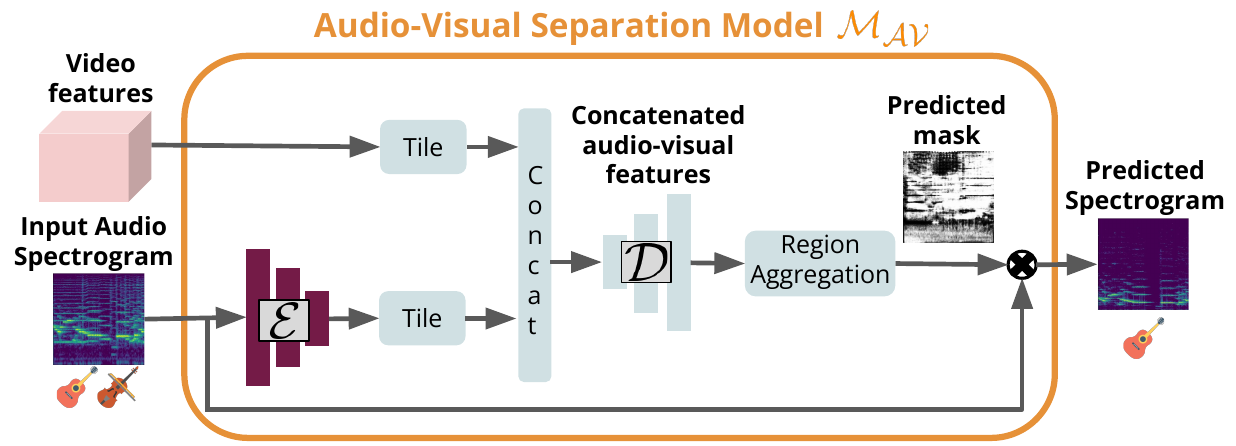}
    \caption{\textbf{Audio-visual separation approach in \modelabb.} We infer a predicted spectrogram mask for each spatiotemporal region and aggregate them to compute a final prediction for the input video.}
    \label{fig:visual_forward_model}
\end{figure}

Given an input video $V$, we begin by using the CLIP visual encoder to extract a set of language-grounded spatiotemporal region representations $f^V \in \mathbb{R}^{T \times H \times W \times D}$. For the $j$-th spatiotemporal region, we tile its visual representation by the factor $H^A W^A$ and concatenate them with the audio bottleneck representations (Figure~\ref{fig:visual_forward_model}) along the channel dimension: $f^{AV}_j = concat(f^A, tile(f^V_j))$, where $f^{AV}_j$ has the dimensions $\mathbb{R}^{H^A \times W^A \times 2D}$. We pass the concatenated representations into the decoder $\mathcal{D}$ consisting of a series of upsampling convolutional layers to generate a real-valued ratio mask: $\hat{M}_j = \mathcal{D}(f^{AV}_j) \in \mathbb{R}^{F \times N}$. To predict the separated audio source, each element of the mask is multiplied with the corresponding location in the input spectrogram: $\hat{A}^S_j = \hat{M}_j \odot A^S$, where $\odot$ denotes the Hadamard product. The mask is then applied to the input spectrogram to predict the audio component corresponding to the video: $\hat{A}^S_j = \hat{M}_j \odot A^S$.

To train the audio U-Net decoder $\mathcal{D}$ to predict spectrogram masks given fused audio-visual and audio-text representation inputs, we use the self-supervised ``mix-and-separate" learning objective since we do not have ground-truth audio source annotations within each training video. Specifically, we synthetically combine the audio of multiple videos and the goal is to use the visual information within each video to separate its corresponding audio waveform. This objective allows us to compute ground-truth ratio spectrogram masks for training without annotations. Next, we describe the generation process of the ground-truth ratio masks for a pair of videos which is also commonly used in prior work \cite{zhao2018sound,gao2019co}; the same process is generalizable to any number of input videos. 
Given a pair of ground-truth audio spectrograms $A_1^S$ and $A_2^S$, we compute their ratio masks as follows:
\begin{equation}
    M_1 = \frac{A_1^S}{A_1^S + A_2^S} \quad\text{and}\quad 
    M_2 = \frac{A_2^S}{A_1^S + A_2^S}
\end{equation} 
We adopt the mask prediction loss \cite{zhao2018sound, gao2019co, chatterjee2021visual} to train the audio U-Net decoder $\mathcal{D}$ for audio separation. Given the pair of predicted masks $\hat{M_1}$ and $\hat{M_2}$, we compute the mask prediction loss as:
\begin{equation}
    \mathcal{L}_{mask} = || \hat{M}_1 - M_1 ||_1 + || \hat{M}_2 - M_2 ||_1
\end{equation}
We note that it is also possible to compute the above-mentioned L1 regression loss using the ground-truth audio spectrograms but prior work \cite{gao2019co,zhao2018sound} has demonstrated it is more numerically stable to use the ratio masks for supervision.

\section{Ablation experiments}
\noindent\textbf{Ablation over region MIL mask prediction vs video-level prediction. } We evaluate the effectiveness of learning to perform source separation at the region level as compared to the video level in Table~\ref{tab:prediction_level}. To perform video-level spectrogram mask prediction, we adopt the same video aggregation function in Sound of Pixels \cite{zhao2018sound}, where the region representations are maxpooled over the channel dimension to compute a final video representation that is passed into the audio U-Net decoder $\mathcal{D}$ (Figure~\ref{fig:visual_forward_model}). We note that our proposed alignment objectives are used in the training of both model variants. We observe that training a model to perform region-level predictions under the MIL formulation results in a significant performance gain over performing video-level predictions, which validates our hypothesis that a model trained to perform video-level predictions may not be able to identify candidate objects that emit sound.

\noindent\textbf{Effect of sharing parameters in U-Net encoder $\mathcal{E}$. } Prior work \cite{gao2019co} learns a separate audio encoder for encoding the predicted audio waveforms to classify them according to discrete audio category labels. Here, we aim to determine the benefit of using shared parameters for our audio encoder component of the U-Net model $\mathcal{E}$ in Table~\ref{tab:shared_audio_encoder_params}. In this case, unlike prior work \cite{gao2019co}, we observe that using a shared audio U-Net encoder to encode the input audio spectrogram for source separation and the predicted spectrogram for the two new losses is integral to improving the final performance of our trained model on audio-visual separation.

\noindent\textbf{Ablation over weights of $\mathcal{L}_{\text{Audio-language}}$ and $\mathcal{L}_{\text{Tri-modal}}$.} We report the results of our ablation over the weights of our proposed audio-language and tri-modal consistency alignment objectives in Table~\ref{tab:solos_sop_loss_weights_ablation}. The results of adding the audio-language consistency loss seem to validate our initial hypothesis that using a lower weight term for this loss is beneficial. As discussed earlier in Section \textcolor{red}{3.1}, this is similar to the multimodal contrastive formulation used for training joint vision-language foundation models such as CLIP and ALIGN. Thus, there is a high probability that we are treating some latent captions as false negatives for each video even though they may contain similar sounding objects. Setting a low weight helps to alleviate this negative consequence. However, we observe that the audio-language consistency loss is still very helpful for improving audio-visual source separation as well as learning a strong transitive alignment between the audio and natural language modality. The reported results also suggest that adding the tri-modal consistency loss also helps to improve performance significantly. In this case, we note that this alignment objective is formulated as a KL divergence minimization problem and does not require negative samples. Consequently, it may not be as important to use a low weight for this term as compared to the audio-language consistency objective.

\begin{table}[h]
\begin{center}
\begin{tabular}{|c|ccc|}
\hline
Prediction & SDR  & SIR   & SAR  \\
\hline 
Video-level & 6.72 & 11.47 & 10.58 \\
Region-level & \textbf{8.58} & \textbf{14.16} & \textbf{12.35} \\
\hline 
\end{tabular}
\end{center}
\caption{\textbf{Comparison between video-level and region-level audio predictions with our trained model on the SOLOS dataset.}}
\label{tab:prediction_level}
\end{table}

\noindent\textbf{Replacing regions with bounding boxes.}
\begin{table}[t]
\centering
    \begin{tabular}{|c|ccc|}
    \hline
      &  SDR $\uparrow$ & SIR $\uparrow$ & SAR $\uparrow$ \\
      \hline
      Regions & 8.58  & 14.16 & 12.35 \\
Boxes & 8.32 & 13.63 & 12.22 \\
     \hline
    \end{tabular}
    \caption{\textbf{Evaluation on SOLOS.} We evaluate our trained model by replacing spatiotemporal region representations with those of detected bounding boxes and their representations.}
    \label{tab:bbox_eval}
\end{table}
To determine if our approach can generalize well to pre-extracted bounding boxes during inference, we evaluate our trained model by replacing spatiotemporal region representations with those of bounding boxes during inference. We encode each bounding box as an image representation separately. Note that this is different from the region representations that are extracted from the modified self-attention operation in CLIP visual encoder (Section~\ref{sec:latent_extraction}). Consequently, our trained models may not generalize well to the different visual representations used during training and inference. We report our results in Table~\ref{tab:bbox_eval}, where we observe that using bounding box representations in our trained models leads to a slightly lower performance in audio-visual separation. 

\begin{figure}[h]
    \centering
    \includegraphics[width=0.5\textwidth]{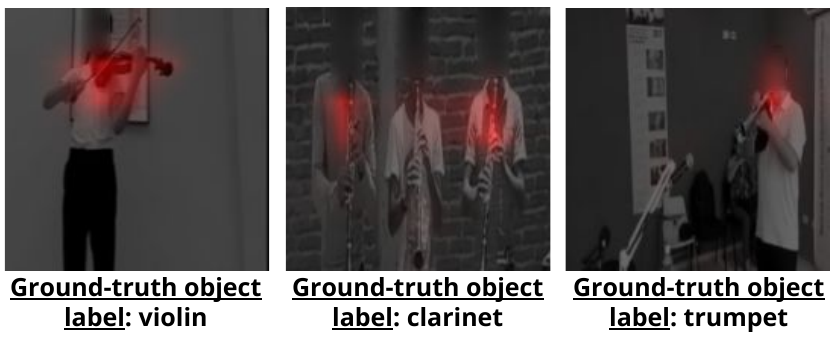}
    \caption{\textbf{Visual attention of latent captions.} We see that the latent captions tend to focus on salient foreground objects.}
\label{fig:latent_concept_attn_map}
\end{figure}

\noindent\textbf{Visualizations of latent captions.} To understand what the latent captions encode, we provide some examples of their attention maps with respect to the video frames in Figure~\ref{fig:latent_concept_attn_map}. Interestingly, we observe that a latent caption is capable of describing multiple instances of the same object in the middle visualization, where it is focusing on all three clarinets.

\begin{table*}[t]
\begin{center}
\begin{tabular}{|c|ccc|ccc|ccc|}
\hline
Shared audio &    & SOLOS &   &    & MUSIC & &    & Audioset & \\
encoder params & SDR $\uparrow$ & SIR $\uparrow$ & SAR $\uparrow$ &  SDR $\uparrow$ &  SIR $\uparrow$ & SAR $\uparrow$ & SDR $\uparrow$ & SIR $\uparrow$ & SAR $\uparrow$\\
\hline 
No & 7.52 & 12.68 & 10.22 & 7.39 & 13.25 & 9.81 & 3.27 & 6.48 & 11.51 \\
Yes & \textbf{8.58} & \textbf{14.16} & \textbf{12.35} & \textbf{8.08} & \textbf{13.97} & \textbf{11.33} & \textbf{11.33} & \textbf{7.62} & \textbf{13.20} \\
\hline 
\end{tabular}
\end{center}
\caption{\textbf{Ablation over using shared parameters for audio U-Net encoder.} We observe that using a common audio encoder $\mathcal{E}$ to encode both mixed and predicted audio inputs for separation and localization, respectively, helps to improve performance on audio-visual separation.}
\label{tab:shared_audio_encoder_params}
\end{table*}

\section{Datasets}
We train and evaluate our proposed \modelabb approach as well as other baselines on the widely-used SOLOS, MUSIC and AudioSet datasets which we describe below.

\noindent\textbf{MUSIC \cite{zhao2018sound}. } The MUSIC dataset consists of videos that are downloaded from YouTube using queries about various musical instruments. It contains approximately 536 and 149 solo and duet videos, respectively. The entire set is comprised of videos containing 11 instrument categories: accordion, acoustic guitar, cello, clarinet, erhu, flute, saxophone, trumpet, tuba, violin and xylophone. Since the original splits of the dataset are not released, we adopt the same splits as \cite{gao2019co}, where the first and second videos in each instrument category are used as validation / test data and the rest are used for training.

\noindent\textbf{SOLOS\cite{montesinos2020solos}. } Similar to the MUSIC dataset, the SOLOS dataset contains 755 videos of musical videos that span 13 instrument categories. These videos are obtained from YouTube where the authors use queries of instruments as well as the `solo' or `auditions' tag. Unlike the MUSIC dataset, the SOLOS dataset does not contain duet videos. 

\noindent\textbf{AudioSet-Unlabeled \cite{gemmeke2017audio}. } AudioSet is a dataset that contains over two million 10 second video clips spanning 632 audio event classes that are sourced from YouTube. Compared to the MUSIC and SOLOS datasets, the audio clips in AudioSet are generally much noisier due to the presence of background sounds. Following prior work \cite{gao2019co}, we filter the video clips according to 15 musical instrument categories and select those from the `unbalanced' split for training and the `balanced' split for validation and testing. 

\section{Implementation details}
We implement our proposed approach using the Pytorch deep learning library \cite{paszke2019pytorch}. Consistent with prior work \cite{zhao2018sound,gao2019co}, we downsample the audio clips to 11 kHz and use a Hann window size of 1022 samples\footnote{While it is common to use powers of 2 as FFT size, we use 1022 as opposed to 1024 to be consistent with previous literature.} and a hop length of 256 samples in the STFT operation. This step results in an audio spectrogram of dimensions 512 x 256, which is re-sampled on a log-frequency scale to compute a final spectrogram of dimensions 256 x 256. We use the CLIP Resnet50 model \cite{radford2021learning} and its language encoder to extract a latent caption for each video as well as encode visual and language representations for audio separation. We set the dimension of the audio U-Net bottleneck features $D$ to be the same as that of CLIP embedding space, which is 1024. We freeze the CLIP encoders during training and train the audio U-Net from scratch using a base learning rate of 4e-3. We train all models for 100 epochs with the SGD optimizer as well as using a linear warmup of 1000 steps and anneal the learning rate using a cosine decay schedule. We train our full model using 4 Quadro 6000 GPUs for approximately 8 days.

\begin{table}[t]
\begin{center}
\begin{tabular}{|c|c|ccc|}
\hline
$\mathcal{L}_{\text{Audio-language}}$ & $\mathcal{L}_{\text{Trimodal}}$ & SDR $\uparrow$ & SIR $\uparrow$  & SAR $\uparrow$ \\
weight & weight &   &   &  \\
\hline 0.0 & 0.0 & 5.47 & 10.55 & 10.95 \\
1e-1 & 0.0 & 6.09 & 11.77 & 10.77 \\
1e-2 & 0.0 & 8.08 & 13.74 & 12.18 \\
1e-3 & 0.0 & 7.45 & 13.40 & 11.11  \\
1.0 & - & 1.24 & 4.97 & 11.27 \\
- & 1e-1 & 8.02 & 13.82 & 11.76 \\
0.0 & 1e-2 & 7.92 & 13.49 & 11.65 \\
0.0 & 1e-3 & 8.10 & 13.84 & 11.79 \\
0.0 & 1.0 & 6.81 & 12.61 & 11.00 \\
1e-3 & 1e-2 & 8.58 & 14.16 & 12.35 \\
\hline 
\end{tabular}
\end{center}
\caption{\textbf{Ablation results over the weights of the audio-language and tri-modal consistency alignment objectives on SOLOs.} We observe that the inclusion of the audio-language and tri-modal consistency alignment objectives is beneficial for audio-visual separation.}
\label{tab:solos_sop_loss_weights_ablation}
\end{table}

\section{Limitations} While we have demonstrated that our proposed \modelabb approach is able to generalize well to free-form natural language queries for source separation, we observe that it is only able to handle visually descriptive adjectives such as \emph{person playing a small trumpet} instead of \emph{a loud trumpet}. We hypothesize that this limitation is due to a higher likelihood of visually descriptive adjectives appearing in the alt text of the pretraining dataset used by CLIP. Additionally, we only focus on separating sounds of different object classes. Our approach does not generalize well to discriminating between sounds from multiple instances of the same class (\cf, Fig {\color{red} 5} middle showing that we can detect the clarinets but not distinguish the different instances). An example of such a challenging task is audio-visual speech separation, where there are two or more people speaking simultaneously and the goal is to separate for the speech for each person. Similar to existing audio-visual speech separation approaches~\cite{ephrat2018looking,rahimi2022reading}, future work can aim to address this limitation by leveraging representations of different instances and additional information in the form of object labels and speech narrations.

\section{Demo video with predicted audio component generations}
We provide a demo video where we evaluate our trained models on random videos in the wild which contain two instruments. The video contains 4 evaluation samples on the task of audio-language source separation in the input videos. Additionally, we also localize the separated audio sources in the corresponding video frames. For the first task, our objective is to separate an audio input based on a natural language query and the goal of the second task is to localize the predicted separated audio in its corresponding video. Note that we use our full \modelabb model that is trained with our proposed audio-language and tri-modal consistency alignment objectives. For each evaluation sample, we provide the following in order:
\begin{enumerate}
    \item Input video with mixed audio input (composed of two different instruments)
    \item Separated audio predicted by the full \modelabb model of the first instrument
    \item Attention heatmap between the first separated audio in (2) and the center frame
    \item Separated audio predicted by the full \modelabb model of the second instrument
    \item Attention heatmap between the second separated audio in (4) and the center frame
\end{enumerate}
We observe that our full \modelabb model, that is trained without ground-truth text annotation or object bounding boxes, is generally able to separate the audio inputs based on natural language queries. 

\end{document}